%% file: neurips_2026.tex
\documentclass{article}

\PassOptionsToPackage{numbers}{natbib}

 \usepackage[preprint]{neurips_2026}

\usepackage[utf8]{inputenc} %
\usepackage[T1]{fontenc}    %
\usepackage{hyperref}       %
\usepackage{url}            %
\usepackage{booktabs}       %
\usepackage{amsfonts}       %
\usepackage{nicefrac}       %
\usepackage{microtype}      %
\usepackage{xcolor}         %
\usepackage{wrapfig}
\usepackage{placeins}

\usepackage{subcaption}

\usepackage{pifont}   %
\usepackage{enumitem} %
\usepackage{graphicx}
\usepackage{amsmath}
\usepackage{multirow}
\definecolor{checkgreen}{RGB}{20, 120, 20}
\definecolor{crossred}{RGB}{150, 0, 0}

\usepackage{xcolor}

\newcount\rainbowidx
\newcount\rainbowtotal
\newcount\rbA %
\newcount\rbB %

\def\rainbowcount#1{\rainbowtotal=0 \rainbowcountloop#1\rainbowstop}
\def\rainbowcountloop#1{%
  \ifx#1\rainbowstop
  \else
    \advance\rainbowtotal by 1
    \expandafter\rainbowcountloop
  \fi
}

\def\rainbowtext#1{%
  \begingroup
  \rainbowcount{#1}%
  \rainbowidx=0
  \rainbowgradloop#1\rainbowstop
  \endgroup
}
\def\rainbowgradloop#1{%
  \ifx#1\rainbowstop
  \else
    \advance\rainbowidx by 1
    \rainbowgradcolor{#1}%
    \expandafter\rainbowgradloop
  \fi
}
\def\rainbowgradcolor#1{%
  \rbA=\rainbowidx
  \multiply\rbA by 100
  \divide\rbA by \rainbowtotal
  \ifnum\rbA>50
    \rbB=\rbA \advance\rbB by -50 \multiply\rbB by 2
    \textcolor{blue!\the\numexpr100-\rbB\relax!purple}{#1}%
  \else
    \rbB=\rbA \multiply\rbB by 2
    \textcolor{teal!\the\numexpr100-\rbB\relax!blue}{#1}%
  \fi
}

\newcommand{\modelname}[1]{PointZero{#1}}
\newcommand{\pretraining}[1]{pre-train#1}

\title{\rainbowtext{PointZero}: 3D Point Track Completion for \\ Learning Transferable 3D Dynamics} %

\author{%
  \makebox[\textwidth][c]{%
  \small
  \begin{tabular}{@{}c@{\hspace{1.5em}}c@{\hspace{1.5em}}c@{\hspace{1.5em}}c@{}}
    \textbf{Bardienus P. Duisterhof} & \textbf{Kaifeng Zhang}\thanks{Equal contribution.} & \textbf{Adam Hung}\footnotemark[1] & \textbf{Bowen Wen} \\
    \mdseries CMU & \mdseries Columbia & \mdseries CMU & \mdseries NVIDIA \\[2ex]
    \textbf{Stan Birchfield} & \textbf{Yunzhu Li} & \textbf{Deva Ramanan} & \textbf{Jeffrey Ichnowski} \\
    \mdseries NVIDIA & \mdseries Columbia & \mdseries CMU & \mdseries CMU \\[1ex]
  \end{tabular}}%
}
\begin{document}
\vspace*{-0.6in}
\maketitle
\vspace{-0.4in}
\begin{center}
\href{https://pointzero-wm.github.io}{\rainbowtext{pointzero-wm.github.io}}
\end{center}
\begin{figure*}[htb!]
\vspace{-13pt}
    \centering
    \includegraphics[width=0.9\linewidth]{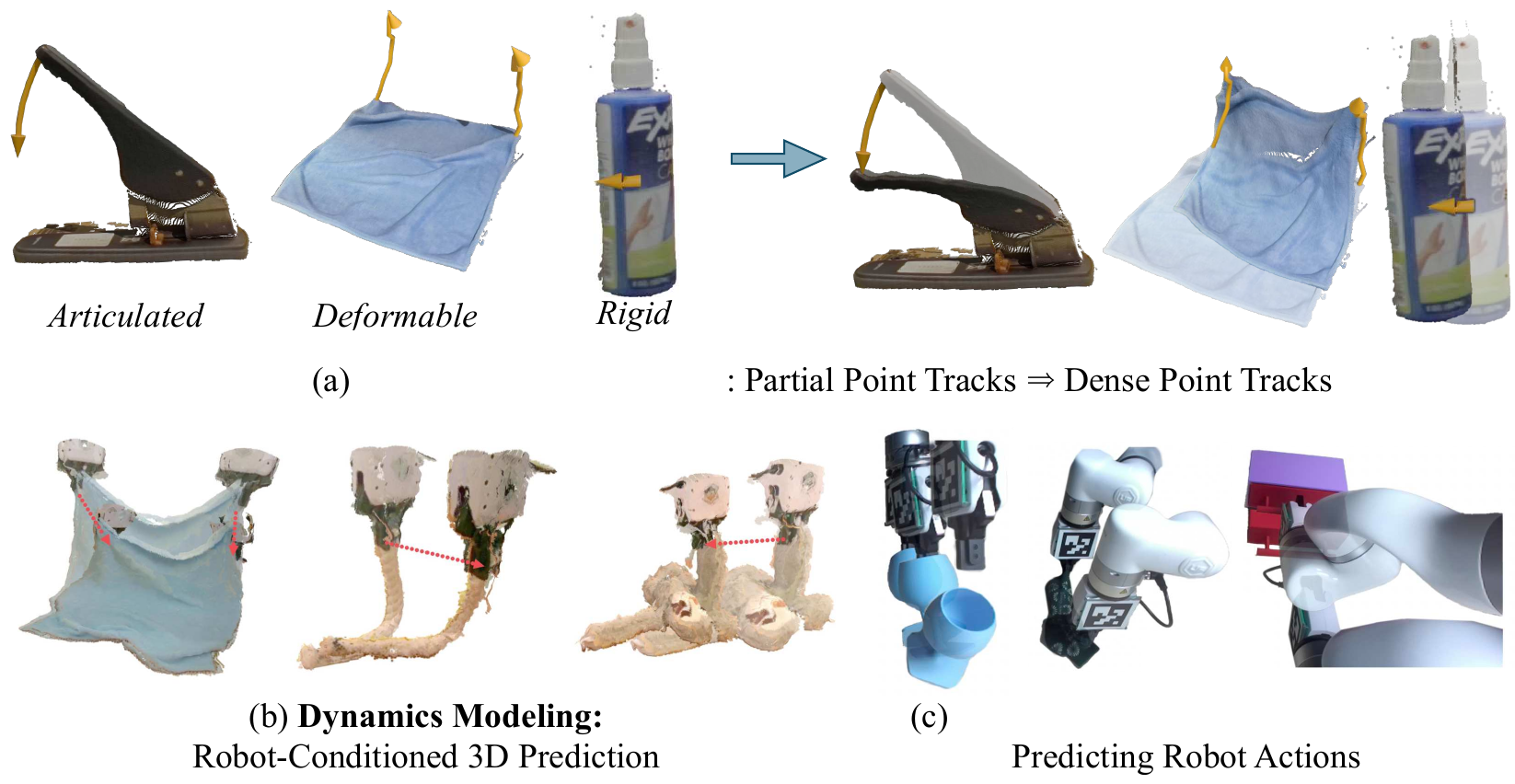}
    \caption{
    \modelname{} proposes \emph{3D point track completion} as a \pretraining{ing} objective for learning rich 3D dynamics priors. Given one RGB-D image and one or a few point trajectories (shown in orange), \modelname{} predicts the future 3D tracks of all observed scene points. 
    \modelname{} is a robot-free pre-training objective for instilling 3D dynamics understanding in world models. 
    Post-trained for dynamics modeling~\cite{zhang2024particle} and robot manipulation~\cite{hung20263pointr}, it outperforms \emph{application-specific} baselines.  
    }
    \label{fig:hero}
    \vspace{-5pt}
\end{figure*}

\begin{abstract}
World models endow perceptual systems with the ability to predict how scenes evolve under interaction. 
They are most beneficial when trained on diverse volumes of data, to instill a rich prior into downstream applications.
Existing methods typically require robot action labels to learn action-conditioned 3D dynamics, which excludes web video data from the training pool. 
We study \emph{3D point track completion} as a \pretraining{ing} objective for learning transferable 3D dynamics without robot data. Given a single RGB-D observation and sparse partial 3D trajectories (tracks), we predict future 3D tracks of all observed points. We show this objective produces a rich 3D dynamics prior, without requiring robot action labels.
We contribute a diverse dataset of 2.9 million synthetic frames spanning deformable, articulated, and rigid objects, and use it to train \modelname{}.
We show that a flexible and expressive transformer, \modelname{}, outperforms prior methods on the same data.
We demonstrate the utility of our pre-training objective by post-training \modelname{} for two downstream applications: (1) action-conditioned 3D dynamics prediction and (2) imitation learning.
When fine-tuned to condition on end-effector pose, \modelname{} outperforms the baselines on the recent PGND 3D dynamics benchmark.
When fine-tuned to predict robot actions and 3D tracks, \modelname{} outperforms or matches the baselines on 6/7 simulated and real-world robot manipulation tasks.
We furthermore evaluate training \modelname{} from scratch to isolate the benefits of our proposed architecture from those of our proposed pre-training objective and dataset.
We release the dataset, checkpoints, and full training recipe.

\end{abstract}

\input{Sections/introduction}
\input{Sections/related_work}

\input{Sections/problem_statement}

\input{Sections/methods}
\input{Sections/results}
\input{Sections/discussion}

\begingroup
\setlength{\bibsep}{4pt plus 1pt minus 0.5pt}
\bibliographystyle{unsrtarxiv}
\bibliography{bibliography}
\endgroup

\appendix

\newpage
\input{Sections/appendix}

\newpage

\end{document}

%% file: Sections/introduction.tex
\section{Introduction}

Humans intuitively anticipate how the world responds to interaction: pulling a drawer, folding a sock, or sliding a cup.
Endowing perception systems with comparable predictive abilities promises to unlock applications across robotics, simulation, and AR/VR/XR. 

\paragraph{Challenges.}
Learning a generic 3D dynamics model is inherently challenging due to the vast \emph{diversity} of physical phenomena, the \emph{underobservability} of many aspects of physics, and the difficulty of characterizing the space of all possible \emph{interactions}.
Despite such challenges, humans readily interact with articulated, rigid, and deformable objects, and can predict plausible dynamics ``zero-shot'' before interacting with a particular object of interest.
Prior methods learn robot-action-conditioned 3D dynamics models by fitting a separate model per scene from hundreds of interactions with the target object~\cite{zhang2024particle,huang2025particleformer}.
Even hundreds of interactions may be insufficient to learn complex dynamics from scratch, motivating pre-training objectives that provide transferable priors over 3D scene dynamics.

\paragraph{Pre-training.}

Large-scale pre-training on video data has led to dramatic improvements in data efficiency and generalization capabilities of robot learning systems~\cite{zheng2026egoscalescalingdexterousmanipulation, wu2023unleashinglargescalevideogenerative}.
However, most such approaches pre-train 2D visual representations or video prediction models, which cannot explicitly represent metric 3D scene motion.

Recent 3D world models have begun scaling dynamics prediction in 3D, but rely on robot interaction data with action labels~\cite{huang2026pointworldscaling3dworld}.  Requiring robot action labels makes data collection considerably more expensive and excludes the vast amount of available web video. This raises the question, \emph{can we learn 3D dynamics priors without action annotations?}

\paragraph{3D Point Track Completion.} We propose 3D point track completion as a robot-free training objective for learning transferable 3D dynamics. 
Given a single RGB-D observation and 3D point trajectories for a few sparse points, we predict the 3D point tracks of all observed points.
Three properties make the objective scalable.
\emph{Robot-free supervision:} The labels and conditioning features are both just 3D point tracks, available from simulation and in principle from any video via point tracking~\cite{karhade2025any4d,karaev2024cotracker3}. No robot- or embodiment-specific annotations are needed.
\emph{Task-agnostic supervision:} As with other dynamics-modeling objectives, training examples need not correspond to successful task completions. Arbitrary play data or free-form interactions with objects can all provide useful supervision.
\emph{Flexible representation:} 3D point tracks can represent arbitrary objects, including those with high-dimensional state such as cloth.

\paragraph{Post-training.}
We study the utility of our pre-training objective by post-training pre-trained models for two practical applications.
First, we adapt the model to condition on robot pose instead of partial point tracks, and train it on scene-specific action-conditioned dynamics modeling using the PGND benchmark~\cite{zhang2024particle}.
Second, we adapt \modelname{} for robot manipulation by adding a small head to predict robot actions and supervise with robot manipulation demonstrations.

\paragraph{Zero-shot generalization.}
We further evaluate whether \modelname{} learns a \emph{generalizable} dynamics prior by testing the model's zero-shot capabilities.
\modelname{}'s architecture consistently outperforms other 3D dynamics-modeling architectures when all methods are trained on our synthetic dataset and evaluated zero-shot on PGND scenes~\cite{zhang2024particle} and on a custom test set of unseen real-world objects.

\paragraph{Contributions.}
\begin{itemize}[nosep,leftmargin=*,labelsep=0.5em]
    \item \textbf{Problem formulation:} We introduce 3D point track completion as a flexible and transferable pre-training objective for learning arbitrary 3D dynamics without requiring action labels.
    \item \textbf{Performant task transfer:} 
    When post-trained for \emph{action-conditioned dynamics modeling} and \emph{robot manipulation}, \modelname{} outperforms strong \emph{application-specific} baselines and training the \modelname{} architecture from scratch.
    This suggests that our pre-trained model learns broadly useful 3D dynamics priors that transfer effectively to robotic applications.

    \item \textbf{Dataset:} We release a diverse 4D synthetic dataset with dense per-point trajectory annotations spanning deformable, articulated, and rigid object interactions. We also release the training and inference code for pre- and post-training, as well as pre-trained checkpoints.
\end{itemize}

%% file: Sections/related_work.tex
\section{Related Work}

\paragraph{Scene-specific Dynamics Models.}
A common problem setting in prior work is to fit a dynamics model to interaction data collected in a specific scene. Methods grounded in physics such as mass-spring systems~\cite{10.1145/2508363.2508406, jiang2025phystwin}, the Finite Element Method (FEM)~\cite{https://doi.org/10.1002/cav.142}, Position-Based Dynamics (PBD)~\cite{flex}, and the Material Point Method (MPM)~\cite{SULSKY1995236, li2023pac} perform well only when state estimation and system identification are successful. %

The difficulty in properly identifying system parameters under partial observability and noise has led researchers to explore learning-based approaches.
Graph-Based Neural Dynamics (GBND)~\cite{li2019learningparticledynamicsmanipulating, sanchez2020learning, zhang2024adaptigraph} has shown great promise in simulating complex, high-dimensional deformable objects.
However, its reliance on manually designed graph topologies and local message passing limits its applicability across a broader range of objects, particularly to rigid and articulated objects where dynamics involve abrupt changes and constrained motion.
Several works explored particle-based predictions without relational edges~\cite{zhang2024particle, whitney2024modeling, huang2025particleformer, tian2025uniclothdiff}
that can learn dynamics models successfully on challenging materials, but come with a limitation that the methods are only trained on single objects or scenes with predefined physical parameter types.
In contrast, \modelname{} combines a simple transformer architecture with a 3D point track completion objective to learn representations that adapt to scene-specific dynamics in post-training, outperforming these baselines on their own benchmarks.

\paragraph{Large-scale 3D Representation Learning.}
Large-scale pre-training has transformed 3D computer vision, enabling models to generalize across diverse scenes, objects, and viewpoints.
DUSt3R~\cite{dust3r_cvpr24} and follow-up works~\cite{wang2025vggt,keetha2026mapanything,duisterhof2024mast3rsfmfullyintegratedsolutionunconstrained} have substantially advanced 3D reconstruction, while other efforts have improved object pose estimation~\cite{wen2024foundationpose}, stereo depth estimation~\cite{wen2025foundationstereo}, and scene completion~\cite{duisterhof2025rayst3r}.

Dynamics models trained at scale have primarily focused on image next-state prediction, either in latent~\cite{zhou2025dinowmworldmodelspretrained} or pixel~\cite{chen2024diffusionforcingnexttokenprediction,li2026hydra0} space (e.g., video generation).
Scalable 3D dynamics learning has remained largely unexplored, owing to the scarcity of large 3D temporal datasets.
Recent work explores language-conditioned 3D point forecasting~\cite{zhang2026molmomotionforecastingpointtrajectories} and pose-trajectory completion~\cite{lei2026world}.

Concurrent to our work, PointWorld~\cite{huang2026pointworldscaling3dworld} studies scalable 3D dynamics prediction and can model scene dynamics without scene-specific interaction data or explicit object geometry priors.
However, it requires calibrated real-world robot interaction data to train, which limits scalability.
\modelname{} instead pre-trains on synthetic point track completion, conditioning on sparse tracks to predict dense trajectories without requiring any robot annotations.

\paragraph{Pre-training for Robotics.}
Prior works investigated \pretraining{ing} visual representations for downstream manipulation policies.
One approach is to learn generic visual features using self-supervised learning methods such as masked autoencoding \cite{MAE, qian20253d} or DINO \cite{caron2021emerging}, and then fine-tune on robotics tasks. Rising in popularity is predicting task-conditioned futures~\cite{hung2026modalityautoregressiveworldactionmodels,yan2026flexpi,ye2026worldactionmodelszeroshot} across modalities such as video, DINO, point tracks and depth. Other methods learn representations through explicit supervision over manipulation-relevant signals, such as video-language alignment~\cite{nair2022r3m}, human actions~\cite{shaw2022videodexlearningdexterityinternet}, or point trajectories~\cite{hung20263pointr, wen2024anypointtrajectorymodelingpolicy}.
These works suggest that structuring representation learning around features relevant to manipulation can improve downstream policy learning.
Our 3D point track completion objective follows a similar principle: by predicting dense 3D point trajectories, the model is encouraged to learn representations that capture task-relevant object geometry, motion, and dynamics.
The closest setting to ours is 3PoinTr \cite{hung20263pointr}, which uses task-specific 3D point track prediction as pre-training for downstream manipulation; we compare against 3PoinTr and other baselines in our imitation-learning experiments.

%% file: Sections/problem_statement.tex
\section{Problem Statement}
\label{sec:problem}

We introduce the pre-training objective of \emph{3D point track completion}.

\textbf{Point cloud observation.} Starting from a single RGB-D observation, we use the RGB image $I \in \mathbb{R}^{H \times W \times 3}$, depth map $D \in \mathbb{R}^{H \times W}$, foreground mask $M \in \{0,1\}^{H \times W}$, and camera intrinsics $K \in \mathbb{R}^{3 \times 3}$ to unproject the masked depth into an observed 3D point cloud $P^{\mathrm{obs}} \in \mathbb{R}^{N_p \times 3}$, where $N_p$ is the number of foreground pixels.

\textbf{Partial point track conditioning.} We condition the model on a small number of partial 3D trajectories: we select $N_a$ conditioning points over $T$ timesteps, represented as $A \in \mathbb{R}^{T \times N_a \times 3}$, where empirically $N_a \in \{1, 2, 3\}$.

\textbf{Prediction.} Given the initial observation $P^{\mathrm{obs}}$, visual features extracted from $I$, and the conditioning $A$, \modelname{} completes point tracks $P \in \mathbb{R}^{T \times N_p \times 3}$, with $P_{1,i}=P_i^{\mathrm{obs}}$ and predicted positions at frames $\tau\in\{2,\ldots,T\}$.

The point-level formulation provides a unified dynamics representation for arbitrary object types.
The pre-training objective can be used or adapted for several applications, including
\textbf{partial point track completion:} the trained model directly predicts dense point tracks, given one or more conditioning point tracks.
\textbf{Post-training for action-conditioned dynamics:} replace the partial point track conditioning with robot end-effector pose conditioning.
\textbf{Post-training for robot manipulation:} attach a lightweight action prediction head to the pre-trained representation and supervise it with observation-action pairs from expert robot demonstrations.

%% file: Sections/methods.tex
\section{Methods}

We propose \modelname{}, a single-transformer architecture for modeling action-conditioned 3D dynamics.
The model combines (i)~point cloud geometry, (ii)~action trajectories, and (iii)~visual features into a single diffusion transformer (DiT) that predicts the target state from noise.

\begin{figure*}[t] %
    \centering
    \includegraphics[width=0.95\linewidth]{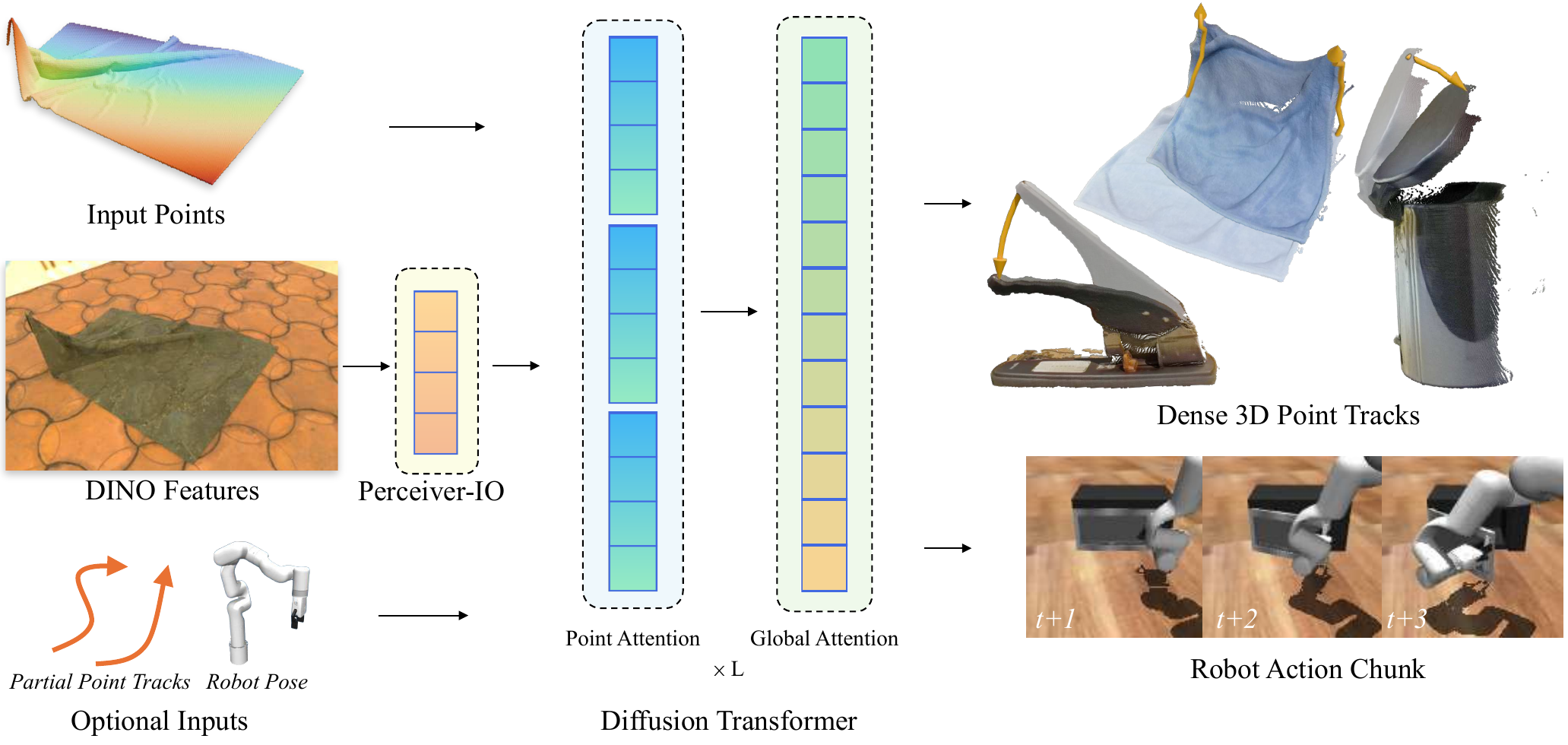}
    \caption{\modelname{} architecture overview. \modelname{} takes in a single RGB-D image and a small number of complete point tracks, and predicts dense 3D point trajectories. 
    \modelname{} can also be fine-tuned to condition on robot pose instead of partial point tracks, or to output robot action chunks in addition to point track predictions.
    \modelname{} combines a Perceiver-IO~\cite{jaegle2021perceiverio} architecture with self-attention and cross-attention to encode diverse dynamics.}
    \label{fig:method}
    \vspace{-10pt}
\end{figure*}

\paragraph{Point tokens.}
Given the observed point cloud $P^{\text{obs}}$, we repeat each initial position over the $T-1$ future frames to form $r_i\in\mathbb{R}^{3(T-1)}$, and embed it using a shared MLP $E_P$, yielding $X_P\in\mathbb{R}^{N_p\times d}$.

\paragraph{Partial point track tokens.}
For each time step $\tau \in \{1,\ldots,T\}$ and trajectory index $j \in \{1,\ldots,N_a\}$, we embed the 3D coordinate $A_{\tau, j} \in \mathbb{R}^3$ using a shared MLP $E_A$.
We inject positional information using a time embedding $e_t(\tau) \in \mathbb{R}^d$, produced by an MLP from the scalar $\tau$, and a trajectory-index embedding $e_a(j) \in \mathbb{R}^d$, looked up from a learned table of $N_a$ vectors.
The resulting action token is $(X_A)_{\tau,j}=E_A(A_{\tau,j})+e_t(\tau)+e_a(j)$, and we flatten these tokens into a single action sequence $X_A \in \mathbb{R}^{(TN_a) \times d}$.

\paragraph{Visual tokens.}
We extract dense image features using DINOv2~\cite{dinov2}, yielding tokens $F_{\text{DINO}}\in\mathbb{R}^{\lfloor H/14 \rfloor \times \lfloor W/14 \rfloor \times d}$.
To reduce memory, we compress them into $N_V$ visual tokens using a Perceiver-IO module~\cite{jaegle2021perceiverio}: starting from learned latent queries $Z\in\mathbb{R}^{N_V\times d}$, we apply $L$ layers of self-attention over $Z$ and cross-attention from $Z$ to $F_{\text{DINO}}$.
This produces $X_V = \mathrm{Perceiver}(Z, F_{\text{DINO}})\in\mathbb{R}^{N_V\times d}$.

\paragraph{Diffusion transformer (DiT).}
We use a diffusion transformer with alternating layers of self-attention and cross-attention.
For each point $i$, the query concatenates the current noisy future trajectory $p_{i,t}\in\mathbb{R}^{3(T-1)}$ with $r_i$, the initial XYZ position repeated $T-1$ times, and projects the result with a shared MLP $E_Q:\mathbb{R}^{6(T-1)}\rightarrow\mathbb{R}^{d}$.
This yields $X_{Q,i}=E_Q(\mathrm{concat}(p_{i,t},r_i))$, with $X_Q\in\mathbb{R}^{N_p\times d}$; a learned diffusion-time embedding is added to each query.

The cross-attention keys and values are formed by concatenating the point, action, and visual embeddings: $X_K = \mathrm{concat}(X_P, X_A, X_V)\in\mathbb{R}^{(N_p+TN_a+N_V)\times d}$.

\subsection{Training Objectives}

We supervise the full trajectory per point to provide a rich supervision signal.
We index the $T=10$ physical frames by $\tau\in\{1,\ldots,T\}$, with the initial observation at $\tau=1$, and define the flattened future trajectory $p_i=\mathrm{flatten}(P_{i,2:T})\in\mathbb{R}^{3(T-1)}$,
where $P_i \in \mathbb{R}^{T \times 3}$ denotes the ground-truth trajectory of point $i$ in $P$.
For all losses below, $\|\cdot\|_w^2$ denotes the coordinate-averaged, temporally weighted squared error defined in Appendix~\ref{app:training_loss}; losses are averaged over points and training examples.

\paragraph{Regression.}
Given the dynamics model $g_{\theta}$ parameterized by $\theta$, we directly regress the trajectory
\begin{equation}
\mathcal{L}_{\mathrm{reg}}
=
\frac{1}{N_p}\sum_{i=1}^{N_p}
\left\lVert
g_{\theta,i}(P^{\text{obs}}, A, I) - p_i
\right\rVert_w^2.
\end{equation}

\begin{figure*}[t] %
    \centering   \includegraphics[width=0.9\linewidth]{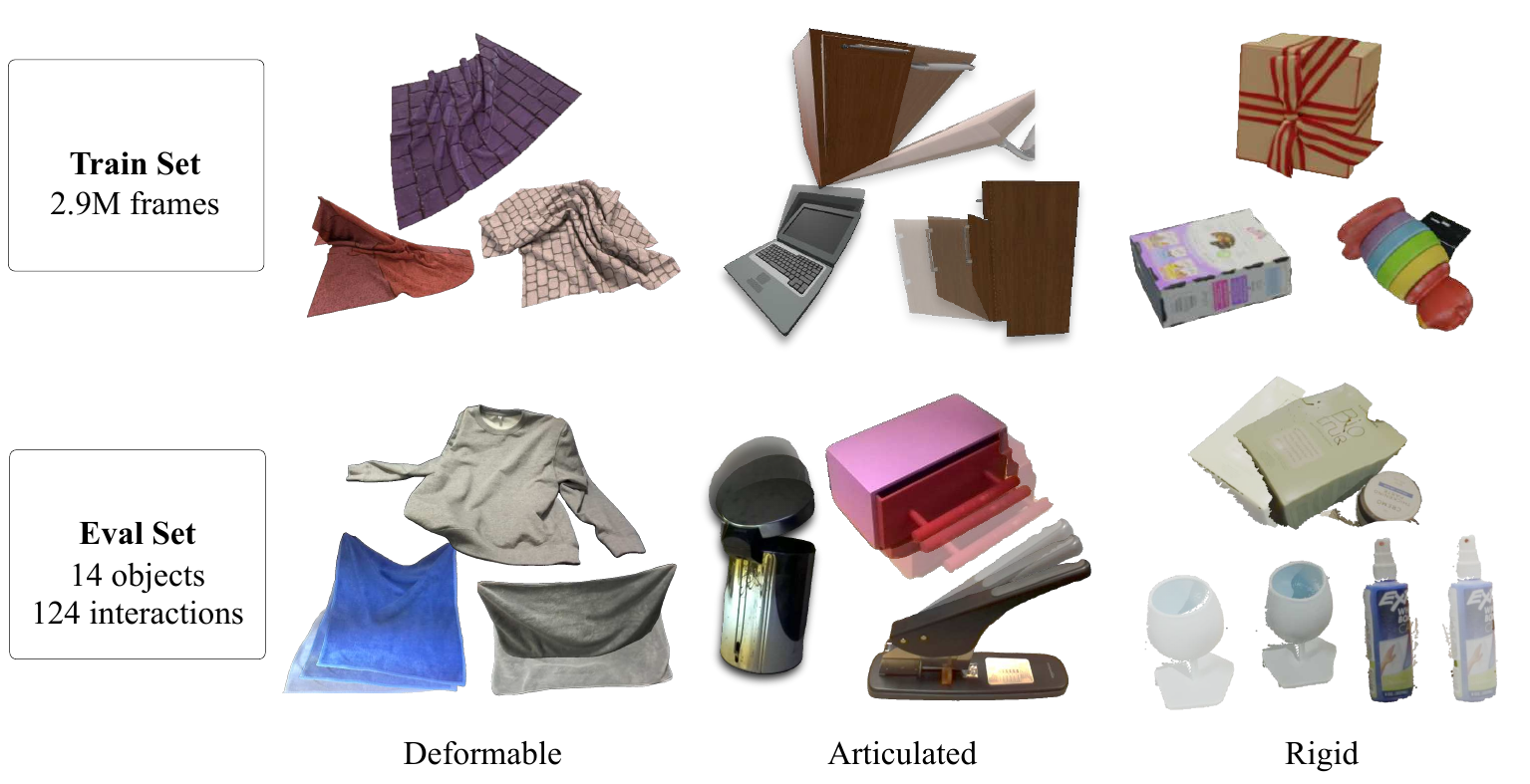}
    \caption{
    We contribute a novel dataset to learn diverse dynamics in simulation and the real world. The dataset contains 2.9 million synthetic image frames with deformable, articulated and rigid objects. We also collect a dataset with 14 objects and 124 interactions in the real world for evaluation of zero-shot sim-to-real transfer. }
    \vspace{-5pt}
\label{fig:datasets}
\end{figure*}

\paragraph{Flow matching (v-prediction).}
For flow matching~\cite{lipman2023flowmatchinggenerativemodeling}, we sample an isotropic Gaussian source point $p^{\text{src}}_i=\varepsilon_i$, with $\varepsilon_i\sim\mathcal{N}(\mathbf{0},0.2^2\mathbf{I}_{3(T-1)})$ in normalized coordinates, and interpolate between source and target as
$p_{i,t}=(1-t)p^{\text{src}}_i+t p_i$ for diffusion time $t \in [0,1]$.
The target velocity is $v_i^\star=(p_i-p_{i,t})/(1-t)=p_i-p_i^{\text{src}}$.
Let $P_t \in \mathbb{R}^{N_p \times (T-1) \times 3}$ be the reshaped tensor corresponding to $\{p_{i,t}\}_{i=1}^{N_p}$.
We train the model to predict $v^\star_i$:
\begin{equation}
\mathcal{L}_{\mathrm{FM}}
=
\frac{1}{N_p}\sum_{i=1}^{N_p}
\left\lVert
g_{\theta,i}(P_t, P^{\text{obs}}, A, I, t) - v^\star_i
\right\rVert_w^2 .
\end{equation}

\paragraph{JiT-style (x-prediction).}

Flow matching and diffusion models for image generation are often applied in learned autoencoder latent spaces, enabling generation in a more compact and smooth representation~\cite{rombach2022high,dao2023flow}.
In contrast, Just image Transformers (JiTs)~\cite{li2026basicsletdenoisinggenerative} perform generation directly in pixel space. JiT argues that natural images lie on a low-dimensional manifold within pixel space, and therefore trains the model to directly predict the denoised image $x_0$.

Similarly, \modelname{} does not use a low-dimensional VAE to encode 3D flow, and we hypothesize that the data also lies on a low-dimensional manifold.
Accordingly, we follow JiT and train the model to perform x-prediction: $g_\theta$ outputs the trajectory $\hat{p}_i = g_\theta(P_t, P^{\text{obs}}, A, I, t)_i$.
Following the JiT training objective, we supervise the implied velocity $\hat{v}_i = (\hat{p}_i-p_{i,t})/(1-t)$ against the target velocity $v^\star_i$:
\begin{equation}
    \mathcal{L}_{\mathrm{JiT}} = \frac{1}{N_p} \sum_{i=1}^{N_p}\left\| \frac{g_{\theta,i}(P_t,P^{\text{obs}},A,I,t)-p_{i,t}}{1-t}-v_i^\star\right\|_w^2.
\end{equation}
Following JiT, we clip the denominator $1-t$ to a minimum of 0.05 when computing both target and predicted velocities during training.

\subsection{Dataset}

A major bottleneck for scaling 3D dynamics learning is the absence of standardized datasets that capture action-conditioned, physically grounded interactions across diverse object types.
Unlike video prediction, 3D dynamics learning requires temporally consistent geometry, explicit object states, and controllable interventions, making data collection particularly costly and fragmented across prior work.
To this end, we contribute a new synthetic dataset spanning deformable, articulated, and rigid objects with motions driven by randomized interactions.
Details regarding our dataset generation are provided in Appendix~\ref{app:datagen}.

\subsection{Post-training Recipes}
\label{sec:posttraining}
We study the utility of 3D point track completion as a \pretraining{ing} objective for downstream tasks.

\paragraph{Post-training on end-effector state.}
For accurate scene-specific dynamics, we replace the conditioning on sparse partial point tracks with conditioning on robot end-effector (EEF) pose.
Concretely, we swap the action tokens $X_A$ for EEF tokens that encode the 6-DoF pose and gripper state at each timestep, embedded by a new MLP $E_{\text{EEF}}$.
We initialize compatible components from the pre-trained model and fine-tune on the target scene.

\paragraph{Imitation-learning post-training.}
For policy learning, we remove partial point track conditioning and add an action head $\pi_\phi$ that predicts an EEF trajectory.
With point-track supervision, we jointly predict EEF and point trajectories. See Section~\ref{sec:posttraining_il} and Appendices~\ref{app:manipulation_recipes} and~\ref{app:imitation_learning} for task-specific recipes.

%% file: Sections/results.tex
\section{Results}

\subsection{Implementation Details}

\paragraph{Normalization and augmentations.}
We normalize each point cloud using the initial-frame centroid and 99$^\text{th}$-percentile radius, and invert this transformation before evaluating errors in physical units.
We then apply random rotations and Gaussian noise to both the point clouds and partial point track conditioning points to enhance robustness to noisy real-world measurements.
We also apply visual augmentations to image observations, including random changes in brightness and contrast, salt-and-pepper noise, and Gaussian noise.

\paragraph{Training.}
We train \modelname{} on a node of 8$\times$~H100 GPUs for approximately 2 days per variant.
Following JiT~\cite{li2026basicsletdenoisinggenerative}, we use logit-normal time sampling and additionally set $t{=}0$ with probability $p_{t=0}=0.2$.
At test time, both flow matching and JiT use four forward-Euler updates on a deterministic logit-normal quantile time grid with parameters $\mu=-3$ and $\sigma=1$.
We adopt a ViT-Base backbone (768 token dimension, 12 heads, 12 layers) with a 3-layer Perceiver-IO encoder and 4 query tokens in the pre-training checkpoints.
Post-training omits DINO visual conditioning for efficiency.

\subsection{Baselines}

We implement several strong existing learning-based dynamics methods, all \emph{trained on our training dataset} with identical augmentations and a prediction horizon of $H=10$.
All methods use the same inputs, conditioning signals, output targets, and supervision; in particular, each baseline is adapted to condition on the same partial point tracks as \modelname{}.

 \noindent\textbf{Graph-Based Neural Dynamics (GBND)~\cite{zhang2024dynamics}:} GBND downsamples the point cloud to 100 particles per scene, connects edges based on top-$k$ nearest neighbors ($k=5$) to form a spatial graph, and applies a GNN to predict next-step per-particle velocities. GBND is rolled out sequentially for longer predictions.
\noindent\textbf{ParticleFormer~\cite{huang2025particleformer}:} ParticleFormer uses the same input as GBND but uses a transformer for particle encoding and predicted velocity decoding.
\noindent\textbf{Particle-Grid Neural Dynamics (PGND)~\cite{zhang2024particle}:} PGND uses PointNet~\cite{qi2017pointnet} and a grid-based representation to predict next-step particle velocities.
\noindent\textbf{Point Transformer v3 (PTv3)~\cite{wu2024ptv3}:} 
PTv3 predicts next-step particle motions and performs sequential prediction in a similar way to PGND, but with a transformer architecture.

\subsection{Evaluation Metrics}

Following prior work~\cite{zhang2024dynamics, huang2025particleformer, zhang2024particle}, we evaluate prediction accuracy using the final-timestep point clouds.
Given ground-truth points $G$ and predicted points $P$, we report the mean squared error (MSE), mean distance error (MDE), the bi-directional Chamfer Distance (CD), and the Earth Mover's Distance (EMD).
Formal definitions of these metrics are provided in Appendix~\ref{app:eval_metrics}.

\subsection{Synthetic Evaluation}
\label{sec:eval_synth}

\input{Tables/results_synth}

We evaluate on held-out samples of our training dataset: approximately 32k scenes across all object categories.
We also provide ablations over model and data size in Appendix~\ref{app:ablations}.

\paragraph{Comparison against baselines.}
Table~\ref{tab:results_synth} shows that every \modelname{} variant outperforms every baseline on every metric and object category. 
This suggests that a flexible but expressive transformer, generating entire motion sequences, is highly performant for learning diverse dynamics. 
\paragraph{Training objectives.}

Generative models are capable of modeling arbitrary distributions given incomplete observations.
We generate ten seeded predictions and report the lowest error independently for each scene and metric. This is a ground-truth oracle.
The generative objectives, FM and JiT, outperform the regression variant, suggesting that they capture a multimodal distribution over plausible point trajectories.
More comprehensive statistics and analysis are available in Appendix~\ref{app:additional_synth_results}.

In our experiments, JiT does not consistently outperform flow matching (\modelname{}-FM) when both models are trained for sufficiently long.

\subsection{Real-World Zero-shot Partial Point Track Completion}
\label{sec:eval_real}

We evaluate zero-shot generalization to real-world objects: methods are trained only on our simulation dataset and tested on novel real-world scenes without fine-tuning.

\paragraph{Data collection.}
We collect 124 interactions across 14 objects:  6 articulated, 5 deformable, and 3 rigid objects, where a human operator manipulates each object.
We obtain action and point-track trajectory labels using FoundationStereo~\cite{wen2025foundationstereo} and CoTracker3~\cite{karaev2024cotracker3}.
To obtain action labels, we manually annotate one human-object contact point per human hand in the first frame and forward-propagate the trajectories.

\paragraph{Analysis.}

\begin{figure*}[t]
    \centering
    \includegraphics[width=0.95\linewidth]{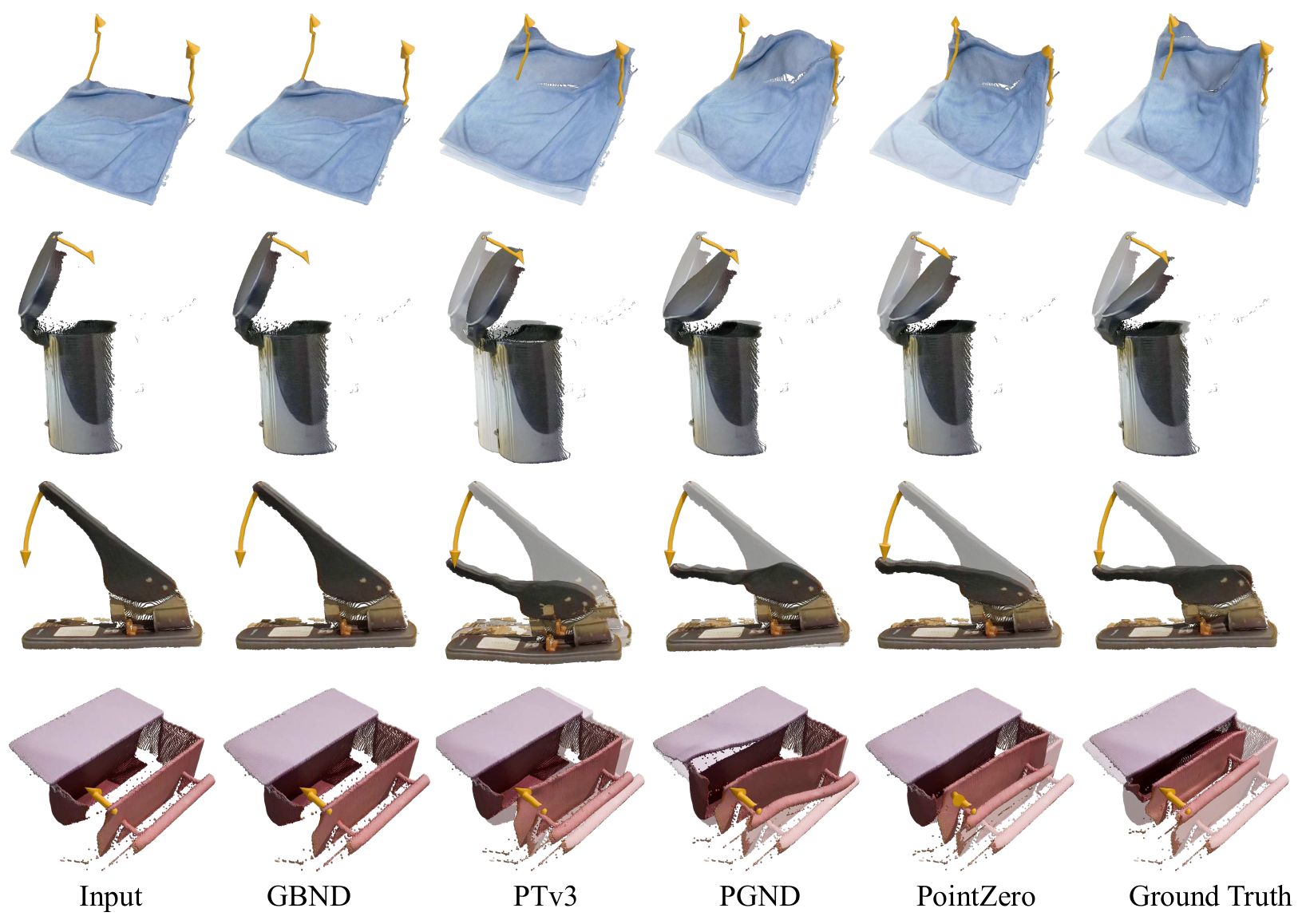}
    \caption{
    Real-world qualitative 3D point track completion evaluation. All methods are trained with identical inputs and conditioning, and with identical data. We find GBND often predicts zero motion, while PTv3 and PGND struggle to recover rigid (articulated) motions.
    }
    \label{fig:qualitative}
    \vspace{-5pt}
\end{figure*}

We display qualitative results in Fig.~\ref{fig:qualitative}.
Compared to baselines, \modelname{} excels at adhering to the input trajectory actions (orange arrows) it is conditioned on.
\modelname{} point track completions are also more faithful to object-specific physical structure, with better preservation of cloth surface area and less deformation of rigid bodies. 
Table~\ref{tab:results_real} shows that both FM and JiT outperform all baselines on 11 of 12 real-world metrics; PTv3 wins on rigid-object MSE.
This suggests that for learning diverse dynamics, simple transformer architectures beat networks with inductive biases such as spatial grids~\cite{zhang2024particle}. 

\subsection{Post-training: Robot-Conditioned 3D Dynamics Prediction}
\label{sec:posttraining_pgnd}
\input{Tables/pgnd_finetune}

3D point track completion is an incomplete approximation of contact-rich real-world dynamics, as is common in robot manipulation.
We post-train \modelname{} (Section~\ref{sec:posttraining}) to predict 3D dynamics conditioned on robot end-effector state, and study the performance of its pre-trained spatial prior.
We evaluate \modelname{} on the six-scene benchmark from PGND~\cite{zhang2024particle} and compare against strong \emph{application-specific} baselines.
For each scene, we post-train a separate instance of \modelname{} using the same training set used to train the PGND baseline. We post-train with the JiT objective and LoRA. We also train \modelname{}-Scratch with the same objective, random initialization, and all parameters trainable.
The results suggest two things: (1) \modelname{} convincingly outperforms state-of-the-art baselines on 4/6 scenes, and (2) pre-training is critical to its strong predictions.

We additionally evaluate \modelname{} in a zero-shot setting on the PGND benchmark.
It achieves competitive performance with scene-specific baselines, \emph{despite never having interacted with the objects.}
We provide these results and additional details in Appendix~\ref{app:pgnd_zeroshot}.

\subsection{Post-training: Imitation Learning for Robot Manipulation}
\label{sec:posttraining_il}
\input{Tables/imitation_learning}

We test whether the trained representation effectively adapts to robot manipulation using the experimental evaluation protocol from 3PoinTr \cite{hung20263pointr}. 
We independently post-train \modelname{} to predict robot actions in three simulation tasks and four real-world tasks. 
For each task, we use 20 expert demonstrations with robot action labels, and 100 additional videos of expert demonstrations without any action labels.
We provide details about the tasks and policy-training setup in Appendix \ref{app:imitation_learning}.

\paragraph{Baselines.}
\textbf{DP3}~\cite{ze20243ddiffusionpolicygeneralizable} and \textbf{Diffusion Policy~}\cite{chi2024diffusionpolicyvisuomotorpolicy} are performant behavior cloning methods that encode observations and then use conditional diffusion models to generate action chunks.
\textbf{3PoinTr}~\cite{hung20263pointr} and \textbf{ATM}~\cite{wen2024anypointtrajectorymodelingpolicy} pre-train point-track prediction models from videos, and then condition policies on the point-track predictions.

\paragraph{Analysis.}
\modelname{} achieves the highest or joint-highest success rate on six of seven manipulation tasks (Table~\ref{tab:imitation_learning}).
Using only 20 action-labeled demonstrations per simulation task and no additional actionless videos, pre-training increases average success from 80.5\% to 88.2\% with auxiliary point-track supervision (Table~\ref{tab:pointzero_20demos}); without it, the pre-trained recipe achieves 80.0\% versus 74.1\% from scratch (Table~\ref{tab:pointzero_no_annotations}).
The latter recipe freezes the pre-trained point-processing stream, whereas Scratch trains the full model (Appendix~\ref{app:manipulation}).
Both comparisons use 1,000 rollouts per simulation task; bold entries mark the better variant.
The results suggest that 3D point track completion as a pre-training objective helps downstream imitation learning applications. In particular, the isolated pre-training experiments in Table~\ref{tab:pointzero_20demos} and Table~\ref{tab:pointzero_no_annotations} suggest 3D point-track completion directly improves performance.

%% file: Tables/results_synth.tex
\begin{table*}[t]
\footnotesize
\centering
\setlength{\tabcolsep}{2pt} %
\renewcommand{\arraystretch}{1.15}

\caption{Dynamics prediction errors on held-out samples of our generated synthetic dataset. 
All baselines were trained on the proposed dataset. 
MDE, CD, and EMD are in centimeters; MSE is in $10^{-2}$\,m$^2$.
\textbf{Bold} marks the best result in each column, and \underline{underlining} indicates the second-best result.
}

\label{tab:results_synth}
\resizebox{\linewidth}{!}{%
\begin{tabular}{@{}l cccc cccc cccc@{}}
\toprule
\multirow{2}{*}{Method}
& \multicolumn{4}{c}{Deformable}
& \multicolumn{4}{c}{Articulated}
& \multicolumn{4}{c}{Rigid} \\
\cmidrule(lr){2-5}\cmidrule(lr){6-9}\cmidrule(lr){10-13}
& MDE $\downarrow$ & MSE $\downarrow$ & CD $\downarrow$ & EMD $\downarrow$ 
  & MDE $\downarrow$ & MSE $\downarrow$ & CD $\downarrow$ & EMD $\downarrow$ 
  & MDE $\downarrow$ & MSE $\downarrow$ & CD $\downarrow$ & EMD $\downarrow$ \\
\midrule
GBND{\scriptsize ~\cite{zhang2024dynamics}} & 17.38 & 6.15 & 7.45 & 16.47 & 11.15 & 6.46 & 7.38 & 11.11 & 155.72 & 759.93 & 93.30 & 154.44 \\
ParticleFormer{\scriptsize ~\cite{huang2025particleformer}} & 17.35 & 6.15 & 7.42 & 16.44 & 10.86 & 6.46 & 7.11 & 10.83 & 155.04 & 760.43 & 92.76 & 153.77 \\
PGND{\scriptsize ~\cite{zhang2024particle}} & 9.01 & 1.36 & 4.10 & 7.86 & 8.70 & 2.13 & 5.40 & 8.29 & 67.58 & 123.79 & 33.56 & 64.02 \\
PTv3{\scriptsize ~\cite{wu2024ptv3}} & 9.20 & 1.59 & 3.91 & 8.28 & 8.25 & 2.31 & 5.25 & 8.00 & 61.97 & 114.77 & 29.87 & 58.57 \\
\modelname{}-FM-mean-10 & 3.59 & 0.25 & 1.55 & 3.18 & 2.40 & 0.36 & 1.77 & 2.34 & 27.26 & 34.25 & 14.17 & 25.41 \\
\modelname{}-FM-oracle-10 & \textbf{2.80} & \textbf{0.15} & \textbf{1.24} & \textbf{2.41} & \underline{1.96} & \underline{0.25} & \underline{1.48} & \underline{1.91} & \underline{19.91} & \underline{18.80} & \underline{10.39} & \underline{18.43} \\
\modelname{}-Regression & 4.08 & 0.32 & 1.74 & 3.63 & 3.85 & 0.66 & 2.57 & 3.68 & 32.14 & 44.58 & 16.43 & 29.78 \\
\modelname{}-JiT-mean-10 & 3.62 & 0.26 & 1.56 & 3.22 & 2.36 & 0.37 & 1.70 & 2.30 & 28.06 & 39.25 & 14.65 & 26.35 \\
\modelname{}-JiT-oracle-10 & \underline{2.85} & \underline{0.16} & \underline{1.26} & \underline{2.47} & \textbf{1.95} & \textbf{0.24} & \textbf{1.44} & \textbf{1.89} & \textbf{19.46} & \textbf{17.89} & \textbf{10.30} & \textbf{17.99} \\
\bottomrule
\end{tabular}%
}
\end{table*}

%% file: Tables/pgnd_finetune.tex
\newcommand{\mde}{MDE$\downarrow$}
\newcommand{\cd}{CD$\downarrow$}
\newcommand{\emd}{EMD$\downarrow$}

\begin{table*}[t]
\footnotesize
\centering
\setlength{\tabcolsep}{2.0pt}
\renewcommand{\arraystretch}{1.08}
\caption{
Dynamics prediction errors on the PGND benchmark~\cite{zhang2024particle} (centimeters).
The PGND method is trained only on scene-specific interactions. 
We evaluate \modelname{} both trained from scratch on the scene-specific interactions (PointZero-Scratch), and also fine-tuned (PointZero-FT) on the scene-specific interactions from the pre-trained checkpoint. 
Both PointZero rows report a per-scene, per-metric best-of-10 oracle (Section~\ref{sec:posttraining_pgnd}).
\textbf{Bold} marks the best result in each column. 
}
\label{tab:pgnd_finetune}
\resizebox{\textwidth}{!}{%
\begin{tabular}{@{}l *{18}{c}@{}}
\toprule
\multirow{2}{*}{Method}
& \multicolumn{3}{c}{Bread}
& \multicolumn{3}{c}{Paperbag}
& \multicolumn{3}{c}{Cloth}
& \multicolumn{3}{c}{Box}
& \multicolumn{3}{c}{Rope}
& \multicolumn{3}{c@{}}{Sloth} \\
\cmidrule(lr){2-4}
\cmidrule(lr){5-7}
\cmidrule(lr){8-10}
\cmidrule(lr){11-13}
\cmidrule(lr){14-16}
\cmidrule(l){17-19}
& \mde & \cd & \emd
& \mde & \cd & \emd
& \mde & \cd & \emd
& \mde & \cd & \emd
& \mde & \cd & \emd
& \mde & \cd & \emd \\
\midrule
GBND~\cite{zhang2024adaptigraph}
& 3.1 & 3.1 & 1.6 & 3.0 & 4.2 & 1.6 & 7.7 & 8.3 & 3.5 & 4.5 & 6.2 & 3.2 & 6.2 & 7.3 & 3.6 & 7.8 & 6.4 & 3.2 \\
PGND~\cite{zhang2024particle}
& 2.0 & 1.8 & 1.0 & \textbf{1.6} & 2.1 & 0.9 & 4.5 & 4.3 & 2.2 & \textbf{2.2} & \textbf{1.5} & 1.6 & 3.9 & 3.8 & 2.1 & 4.3 & 3.3 & 1.7 \\
\modelname{}-Scratch 
&  4.2 & 3.9 & 1.9 & 4.2 & 4.4 & 1.9 & 5.6 & 4.7 & 2.1 & 3.7 & 3.9 & 2.1 & 9.3 & 10.8 & 5.0 & 7.3
   & 5.9 & 2.8 \\
\modelname{}-FT
& \textbf{1.5} & \textbf{1.1} & \textbf{0.6} & \textbf{1.6} & \textbf{2.0} & \textbf{0.8} & \textbf{4.0} & \textbf{3.4} & \textbf{1.5} & 2.4 & 2.7 & \textbf{1.4} & \textbf{3.3} & \textbf{3.3} & \textbf{1.6} & \textbf{3.9} & \textbf{3.0} & \textbf{1.5} \\
\bottomrule
\end{tabular}%
}
\end{table*}

%% file: Tables/imitation_learning.tex
\begin{table}[t]
\centering
\captionsetup{font=small,skip=4pt,position=top}
\begin{minipage}[t]{0.51\textwidth}
\vspace{0pt}
\caption{Imitation learning success (\%). \modelname{} uses 20 labeled demos and 100 actionless videos per task.}
\label{tab:imitation_learning}
\centering
\footnotesize
\setlength{\tabcolsep}{2.5pt}
\renewcommand{\arraystretch}{1.05}
\begin{tabular*}{\linewidth}{@{\extracolsep{\fill}}lrrrrr@{}}
\toprule
Task & 3PoinTr & DP3 & DP & ATM & \modelname{} \\
\midrule
\multicolumn{6}{@{}l}{\textit{Simulation}} \\
Blockstack & 90.9 & 44.4 & 45.4 & 40.0 & \textbf{99.8} \\
Microwave  & 80.8 & 31.4 & 32.0 & 30.4 & \textbf{93.1} \\
Glass      & 95.2 & 74.9 & 57.5 & 3.9  & \textbf{95.9} \\
\midrule
\multicolumn{6}{@{}l}{\textit{Real-world}} \\
Drawer & 90.0 & 70.0 & -- & 30.0 & \textbf{100.0} \\
Cup    & \textbf{100.0} & 80.0 & -- & 30.0 & \textbf{100.0} \\
Paper  & \textbf{90.0} & 10.0 & -- & 0.0 & 70.0 \\
Sock   & \textbf{90.0} & 20.0 & -- & 0.0 & \textbf{90.0} \\
\bottomrule
\end{tabular*}
\end{minipage}\hfill
\begin{minipage}[t]{0.46\textwidth}
\vspace{0pt}
\input{Tables/imitation_learning_ablations}
\end{minipage}
\end{table}

%% file: Tables/imitation_learning_ablations.tex
\caption{Pre-training \textbf{with} downstream track supervision. Success (\%); 20 demos, no extra videos.}
\label{tab:pointzero_20demos}
\begingroup
\centering
\footnotesize
\setlength{\tabcolsep}{2pt}
\renewcommand{\arraystretch}{1.05}
\begin{tabular*}{\linewidth}{@{\extracolsep{\fill}}lrrrr@{}}
\toprule
Recipe & Blockstack & Microwave & Glass & Avg. \\
\midrule
Scratch    & \textbf{93.5} & 81.4 & 66.6 & 80.5 \\
Pretrained & 93.2 & \textbf{91.9} & \textbf{79.5} & \textbf{88.2} \\
\bottomrule
\end{tabular*}
\par
\endgroup

\vspace{2pt}
\caption{Pre-training \textbf{without} downstream track supervision. Success (\%); 20 demos, no extra videos.}
\label{tab:pointzero_no_annotations}
\begingroup
\centering
\footnotesize
\setlength{\tabcolsep}{2pt}
\renewcommand{\arraystretch}{1.05}
\begin{tabular*}{\linewidth}{@{\extracolsep{\fill}}lrrrr@{}}
\toprule
Recipe & Blockstack & Microwave & Glass & Avg. \\
\midrule
Scratch    & 83.8 & 89.5 & 49.1 & 74.1 \\
Pretrained & \textbf{85.2} & \textbf{90.1} & \textbf{64.6} & \textbf{80.0} \\
\bottomrule
\end{tabular*}
\par
\endgroup

%% file: Sections/discussion.tex
\section{Conclusion}

We present \modelname{}, a 3D dynamics model trained to do \emph{3D point track completion}: given a single RGB-D observation and a small number of sparse point trajectories, the model predicts dense future 3D tracks for all observed points. This objective provides a scalable way to learn dynamics priors without requiring embodiment-specific annotations, using a metric point representation that can represent deformable, articulated, and rigid objects. 

Across synthetic and real-world evaluations, \modelname{} substantially improves dense point-track completion over adapted 3D dynamics baselines. More importantly, the learned representation transfers effectively to downstream tasks: after post-training, \modelname{} improves scene-specific action-conditioned dynamics prediction and achieves strong performance on robot manipulation tasks. These results suggest that 3D point track completion can serve as a useful bridge between scalable robot-free dynamics pre-training and downstream embodied intelligence applications.

\paragraph{Limitations and Future Work.}
\modelname{} remains limited by the coverage and realism of its pre-training data. Although our synthetic dataset spans several object classes, it does not capture the full diversity of real-world materials, contact-rich hand-object interactions, cluttered scenes, or long-horizon dynamics. Future work should augment simulation with pseudo-annotations from real-world videos using 3D reconstruction and point tracking methods. \modelname{} also uses relatively simple conditioning signals: partial point tracks during pre-training and end-effector pose during action-conditioned post-training. Richer interaction representations, such as contact locations and forces, may further improve accuracy. Finally, our downstream evaluations remain task-specific and relatively small-scale; future work should study multi-scene and multi-task transfer across a broader range of applications.

\paragraph{Acknowledgments}
This material is based upon work supported by the National Science Foundation Graduate Research Fellowship Program under Grant Nos. DGE2140739 and DGE2631988. Any opinions, findings, and conclusions or recommendations expressed in this material are those of the author(s) and do not necessarily reflect the views of the National Science Foundation.

This work used Bridges-2 at Pittsburgh Supercomputing Center through allocation CIS250005p.

%% file: Sections/appendix.tex
\section{Appendix}

\subsection{Training Loss Weighting}
\label{app:training_loss}
We give more weight to later timesteps to emphasize longer-horizon predictions, where uncertainty is greater. For future frames $\tau\in\{2,\ldots,T\}$, we use $\|u_i\|_w^2=\frac{1}{3(T-1)}\sum_{\tau=2}^{T}[0.1+0.9((\tau-1)/T)^2]\|u_{i,\tau}\|_2^2$.

\subsection{Dynamics Prediction Evaluation Metrics}
\label{app:eval_metrics}
The definitions below apply to PointZero's corresponding, equal-cardinality final-frame point sets; $\Pi(P,G)$ denotes permutation matrices.

Mean Squared Error (MSE):

\begin{equation}
    \mathcal{L}_{\text{MSE}}(P, G) = \frac{1}{N_p} \sum_{i=1}^{N_p} \| P_i - G_i \|_2^2.
    \label{eq:mse_deriv}
\end{equation}

Mean Distance Error (MDE):

\begin{equation}
    \mathcal{L}_{\text{MDE}}(P, G) = \frac{1}{N_p} \sum_{i=1}^{N_p} \| P_i - G_i \|_2.
    \label{eq:mde_deriv}
\end{equation}

Bi-Directional Chamfer Distance (CD):

\begin{equation}
\mathcal{L}_{\text{CD}}(P, G) = \frac{1}{2N_p} \sum_{p \in P} \min_{g \in G} \| p - g \|_2 + \frac{1}{2N_p} \sum_{g \in G} \min_{p \in P} \| g - p \|_2.
\label{eq:cd_deriv}
\end{equation}

Earth Mover's Distance (EMD):

\begin{equation}
\mathcal{L}_{\text{EMD}}(P, G) = \min_{\pi \in \Pi(P, G)} \frac{1}{N_p} \sum_{i=1}^{N_p} \sum_{j=1}^{N_p} \pi_{ij} \lVert P_i - G_j \rVert_2 .
\label{eq:emd_deriv}
\end{equation}
For the PGND benchmark, EMD is computed using more points than the other metrics, which can yield lower reported distances.

\subsection{Dataset Generation Details}
\label{app:datagen}

Our dataset comprises deformable, articulated, and rigid object interactions. For all datasets, we randomize camera location and intrinsics and use a single-view RGB-D observation.
This models the partial visibility and various viewpoints encountered in the real world.

\paragraph{Deformables. } We follow prior work~\cite{lips2024learningkeypointsroboticcloth} to achieve procedural cloth generation.
\begin{itemize}[nosep,leftmargin=*,labelsep=0.5em]
    \item \textbf{Mesh Generation}: We generate \emph{towels}, \emph{t-shirts} and \emph{shorts} meshes. We randomize both the scale and the proportions.
    \item \textbf{Physics}: We use NVIDIA FleX~\cite{flex} to model deformable physics. We apply one or two action trajectories to the cloth, modeling interactions such as \emph{fold}, \emph{lift}, \emph{drop}, or \emph{push}. Concretely, given constraints on location of action points, we solve for the location of all other points that would produce static equilibrium. To enhance motion diversity, we randomize physics parameters such as stiffness and drag.
    \item \textbf{Appearance}: We render the scenes with Blender~\cite{blender}, with random cloth and background textures from PolyHaven~\cite{polyhaven2026}.
\end{itemize}

\paragraph{Articulated Objects}
We use the PartNet-Mobility dataset~\cite{Xiang_2020_CVPR} to retrieve articulated objects, and place them in Genesis~\cite{Genesis} for simulating motion.
We randomly pick a revolute or prismatic joint to articulate, and set a random configuration
in its viable range to produce static equilibrium.
This alone may result in an invisible articulation, or very small motions.
To address this issue, we reject samples where the 2D optical flow of the sequence is small using privileged information in simulation.

\paragraph{Rigid Objects}
Finally, we aim to instill an understanding of rigid body dynamics and collision into \modelname{}.
We modify the Kubric engine~\cite{greff2021kubric} to include ground-truth 3D trajectories and fewer objects (up to 3 instead of 23 objects).

\paragraph{Sampling Action Trajectories}

For partial point trajectory completion training, we automate the extraction of partial point trajectories.
For deformable objects, we save the action trajectories that were used as kinematic constraints in NVIDIA FleX~\cite{flex} to be used as training input.
For articulated and rigid bodies, we pick 3D point tracks from the $k$ largest displacements.

\paragraph{Dataset Mixing.}
The recorded training weights are 0.33 each for shorts, T-shirts, and towels, 1.0 for articulated objects, and 0.33 for rigid objects, giving category probabilities of approximately 42.67\%, 43.10\%, and 14.22\%, respectively.
This gives an approximate category ratio of $3:3:1$, reflecting our emphasis on learning more complex non-rigid and articulated dynamics while still exposing the model to rigid-body motion and collisions.

\subsection{Additional Manipulation Comparisons}
\label{app:manipulation}
\label{app:manipulation_recipes}
Tables~\ref{tab:pointzero_20demos} and~\ref{tab:pointzero_no_annotations} report the pre-training ablations using only 20 demonstrations per simulation task, with no additional actionless videos.
In Table~\ref{tab:pointzero_20demos}, PointZero initializes from JiT pre-training and scratch initializes randomly; both train all parameters with point-trajectory regression and an $\ell_{2,1}$ action loss, using a separate action branch with learned queries. In Table~\ref{tab:pointzero_no_annotations}, both use only the action loss; PointZero keeps its pre-trained point-processing stream fixed while training the remaining parameters, whereas scratch trains the full model.
With downstream track supervision, pre-training improves two tasks and the average (80.5\% to 88.2\%), while Blockstack changes from 93.5\% to 93.2\%.
Under this protocol, Diffusion Policy~\cite{chi2024diffusionpolicyvisuomotorpolicy} and DP3~\cite{ze20243ddiffusionpolicygeneralizable} achieve average success rates of 45.0\% and 50.2\%, respectively; their per-task results are listed in Table~\ref{tab:imitation_learning}.
Without downstream point-track supervision, pre-trained \modelname{} outperforms its scratch variant on all three tasks, with the largest gap on Glass (64.6\% vs.\ 49.1\%).
DP3 outperforms \modelname{} without downstream track supervision on Glass (74.9\% vs.\ 64.6\%).

\subsection{Ablations}
\label{app:ablations}
\input{Tables/ablations}
\input{Tables/ablation_size}

We study the impact of the training data fraction and of model size.

\paragraph{Training data fraction.}
Table~\ref{tab:ablations} reports performance on the articulated split of the real-world set as the pre-training dataset is reduced, with training epochs scaled inversely to hold the optimization budget comparable.
Performance degrades gracefully: reducing the data to 10\% increases MDE from 2.2 to 3.1\,cm, and further reduction to 5\% yields only a small additional drop.
This suggests the pre-training objective extracts useful dynamics priors even from a fraction of the dataset, while the full dataset remains clearly beneficial.

\paragraph{Model size.}
We additionally train a Small variant from scratch (384 token dimensions, 6 heads, 6 layers) with otherwise identical inputs and training data.
Table~\ref{tab:ablation_size} reports performance in the same setting: shrinking the model degrades all metrics substantially, suggesting that further scaling the Base model could yield additional gains.

\subsection{Additional Synthetic Data Point Dynamics Completion Results}
\label{app:additional_synth_results}

In Table~\ref{tab:results_synth_app}, we expand Table~\ref{tab:results_synth} with additional evaluations of \modelname{}-JiT and \modelname{}-FM.
For each sample, we generate ten predictions using different random noise initializations.
The ``-oracle-10'' rows select the lowest error separately for each scene and metric; ``-first'' uses the first prediction, and ``-mean-10'' averages scalar errors over all ten predictions.
For ``-medoid-10'', each corresponding point's final position is selected from its ten sampled positions to minimize the summed Euclidean distance to the other nine, without using ground truth. Different points may come from different samples.
The ``-pooled-scalar-median-10'' rows take each metric's median over scene--prediction pairs within each dataset, then average dataset medians. These distribution summaries are not directly comparable to mean-error rows.

\modelname{}-JiT-first outperforms \modelname{}-Regression on 11 of 12 metrics; the exception is rigid-object MSE (48.627 versus 44.576 in $10^{-2}$\,m$^2$).
Both per-metric best-of-10 oracles outperform regression on all 12 metrics, showing that sampling can produce candidates with lower reconstruction error.

\input{Tables/results_synth_app}

\subsection{Real-World Dynamics Prediction Quantitative Results}
\label{app:real_world_quant}

In Table~\ref{tab:results_real}, we report point track completion error metrics for \modelname{} and baselines on our real-world dataset.
All methods are trained on the same synthetic dataset and evaluated zero-shot on the unseen real-world objects.
Both \modelname{}-FM and \modelname{}-JiT outperform all baselines on 11 of 12 metrics; PTv3 achieves the lowest rigid-object MSE (11.604\,cm$^2$, versus 12.376\,cm$^2$ for \modelname{}-JiT).

\input{Tables/results_real}

\subsection{PGND Zero-shot Evaluation}
\label{app:pgnd_zeroshot}

We train \modelname{} and several 3D dynamics architecture baselines on point track completion on the \modelname{} synthetic dataset. 
Then, we evaluate the zero-shot performance on the PGND benchmark.
To enable zero-shot predictions without fine-tuning the models to condition on robot pose, we extract a single object-point trajectory per PGND sample, and condition on this track.

Results are shown in Table~\ref{tab:pgnd_zeroshot}.
\modelname{}-FM-ZS achieves an average MDE of approximately 3.8\,cm across the six scenes, compared with 5.2\,cm for the strongest baseline, PTv3-ZS, a reduction of approximately 26\%.

\input{Tables/pgnd_zeroshot}

\subsection{Imitation Learning Experimental Setup}
\label{app:imitation_learning}

\begin{figure}[!htbp]
    \centering
    \includegraphics[width=0.75\linewidth]{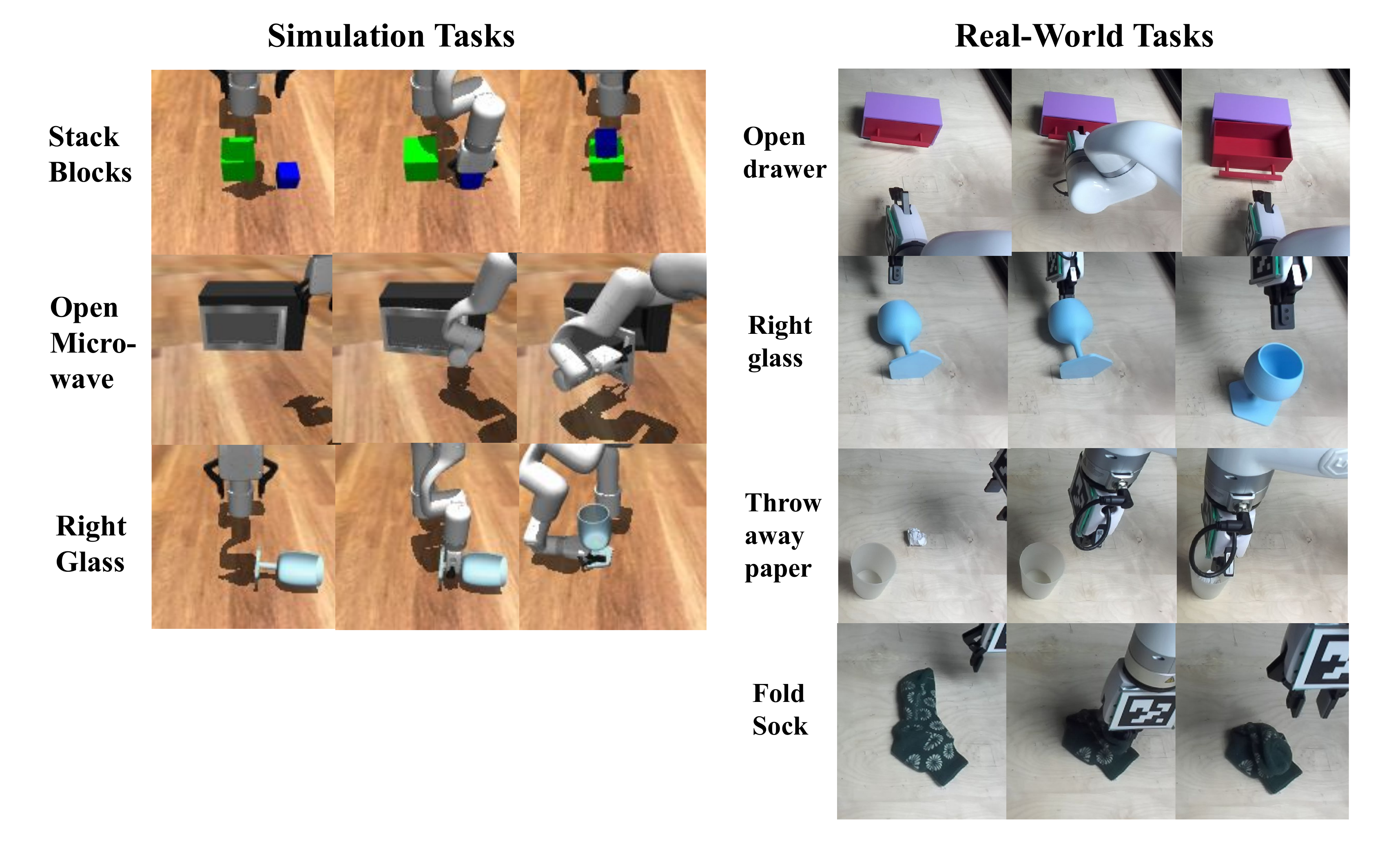}
    \caption{
    Visualizations of the simulation and real-world tasks used for imitation learning evaluation.
    }
    \label{fig:3pointr_tasks}
\end{figure}

Following 3PoinTr~\cite{hung20263pointr}, we evaluate on three simulation tasks: (1) stacking blocks, (2) opening a microwave, and (3) righting a fallen glass.
We also evaluate on four real-world tasks: (1) opening a drawer, (2) righting a fallen glass, (3) picking up a crumpled piece of paper and placing it in the trash, and (4) folding a sock in half.
The position and orientation of the objects are varied across trials, and the same initial configurations are used across all methods.
Figure~\ref{fig:3pointr_tasks} shows images of all the tasks.

%% file: Tables/ablations.tex
\begin{table}[!htbp]
\footnotesize
\centering
\caption{Data-scaling ablation on the articulated split of the real-world dataset:
performance degrades gracefully as the pre-training dataset is reduced.
Error values are in centimeters; MSE is in cm$^2$.
Training epochs are scaled inversely with the data fraction (250 epochs at 10\%, 500 at 5\%).
\textbf{Bold} marks the best result for each metric.
}
\label{tab:ablations}
\renewcommand{\arraystretch}{1.15}
\setlength{\tabcolsep}{3.5pt}
\begin{tabular}{ccccc}
\toprule
Data Fraction & MDE$\downarrow$ & MSE$\downarrow$ & CD$\downarrow$ & EMD$\downarrow$ \\
\midrule
100\% & \textbf{2.2} & \textbf{12.1} & \textbf{1.4} & \textbf{2.0} \\
10\%  & 3.1 & 49.6 & 2.0 & 2.9 \\
5\%   & 3.3 & 64.3 & 2.0 & 3.1 \\
\bottomrule
\end{tabular}
\end{table}

%% file: Tables/ablation_size.tex
\begin{table}[!htbp]
\footnotesize
\centering
\caption{Model-size ablation on the articulated split of the real-world dataset.
Small uses 384 token dimensions, 6 heads, and
6 layers (45M parameters) versus the Base configuration (768 dimensions, 12 heads,
12 layers). Error values are in centimeters; MSE is in cm$^2$.
\textbf{Bold} marks the best result for each metric.
}
\label{tab:ablation_size}
\renewcommand{\arraystretch}{1.15}
\setlength{\tabcolsep}{3.5pt}
\begin{tabular}{ccccc}
\toprule
Size & MDE$\downarrow$ & MSE$\downarrow$ & CD$\downarrow$ & EMD$\downarrow$ \\
\midrule
Base  & \textbf{2.2} & \textbf{12.1} & \textbf{1.4} & \textbf{2.0} \\
Small & 3.1 & 37.4 & 1.8 & 3.0 \\
\bottomrule
\end{tabular}
\end{table}

%% file: Tables/results_synth_app.tex
\begin{table}[!htbp]
    \footnotesize
    \centering
    \setlength{\tabcolsep}{3pt} %
    \renewcommand{\arraystretch}{1.15}
    \caption{Dynamics prediction errors on held-out samples of our generated synthetic dataset.
    All baselines were trained on the proposed dataset. 
    MDE, CD, and EMD are in centimeters; MSE is in $10^{-2}$\,m$^2$.
    First-sample, mean, pointwise medoid, per-metric oracle, and pooled scalar median statistics use ten seeded predictions and are defined in Appendix~\ref{app:additional_synth_results}. Pooled medians summarize a different statistic from mean errors; no cross-statistic ranking is applied.
    }
    \label{tab:results_synth_app}
    \resizebox{\linewidth}{!}{%
    \begin{tabular}{l cccc cccc cccc}
    \toprule
    \multirow{2}{*}{Method}
    & \multicolumn{4}{c}{Deformable}
    & \multicolumn{4}{c}{Articulated}
    & \multicolumn{4}{c}{Rigid} \\
    \cmidrule(lr){2-5}\cmidrule(lr){6-9}\cmidrule(lr){10-13}
    & MDE $\downarrow$ & MSE $\downarrow$ & CD $\downarrow$ & EMD $\downarrow$
      & MDE $\downarrow$ & MSE $\downarrow$ & CD $\downarrow$ & EMD $\downarrow$
      & MDE $\downarrow$ & MSE $\downarrow$ & CD $\downarrow$ & EMD $\downarrow$ \\
    \midrule
    GBND{\scriptsize ~\cite{zhang2024dynamics}} & 17.38 & 6.15 & 7.45 & 16.47 & 11.15 & 6.46 & 7.38 & 11.11 & 155.72 & 759.93 & 93.30 & 154.44 \\
    ParticleFormer{\scriptsize ~\cite{huang2025particleformer}} & 17.35 & 6.15 & 7.42 & 16.44 & 10.86 & 6.46 & 7.11 & 10.83 & 155.04 & 760.43 & 92.76 & 153.77 \\
    PGND{\scriptsize ~\cite{zhang2024particle}} & 9.01 & 1.36 & 4.10 & 7.86 & 8.70 & 2.13 & 5.40 & 8.29 & 67.58 & 123.79 & 33.56 & 64.02 \\
    PTv3{\scriptsize ~\cite{wu2024ptv3}} & 9.20 & 1.59 & 3.91 & 8.28 & 8.25 & 2.31 & 5.25 & 8.00 & 61.97 & 114.77 & 29.87 & 58.57 \\
    \modelname{}-Regression & 4.08 & 0.32 & 1.74 & 3.63 & 3.85 & 0.66 & 2.57 & 3.68 & 32.14 & 44.58 & 16.43 & 29.78 \\
    \modelname{}-FM-first & 3.60 & 0.25 & 1.56 & 3.17 & 2.39 & 0.35 & 1.77 & 2.33 & 28.97 & 38.79 & 15.16 & 27.20 \\
    \modelname{}-FM-mean-10 & 3.59 & 0.25 & 1.55 & 3.18 & 2.40 & 0.36 & 1.77 & 2.34 & 27.26 & 34.25 & 14.17 & 25.41 \\
    \modelname{}-FM-pooled-scalar-median-10 & 3.21 & 0.14 & 1.30 & 2.80 & 1.75 & 0.05 & 1.47 & 1.74 & 11.65 & 1.85 & 7.03 & 11.27 \\
    \modelname{}-FM-medoid-10 & 3.26 & 0.21 & 1.40 & 2.87 & 2.23 & 0.33 & 1.65 & 2.18 & 24.46 & 28.57 & 12.59 & 22.74 \\
    \modelname{}-FM-oracle-10 & 2.80 & 0.15 & 1.24 & 2.41 & 1.96 & 0.25 & 1.48 & 1.91 & 19.91 & 18.80 & 10.39 & 18.43 \\
    \modelname{}-JiT-first & 3.65 & 0.26 & 1.59 & 3.24 & 2.36 & 0.36 & 1.70 & 2.29 & 30.26 & 48.63 & 16.03 & 28.67 \\
    \modelname{}-JiT-mean-10 & 3.62 & 0.26 & 1.56 & 3.22 & 2.36 & 0.37 & 1.70 & 2.30 & 28.06 & 39.25 & 14.65 & 26.35 \\
    \modelname{}-JiT-pooled-scalar-median-10 & 3.27 & 0.15 & 1.31 & 2.87 & 1.65 & 0.06 & 1.37 & 1.64 & 11.23 & 1.86 & 6.96 & 11.01 \\
    \modelname{}-JiT-medoid-10 & 3.32 & 0.22 & 1.42 & 2.92 & 2.23 & 0.33 & 1.61 & 2.17 & 24.93 & 33.08 & 12.81 & 23.32 \\
    \modelname{}-JiT-oracle-10 & 2.85 & 0.16 & 1.26 & 2.47 & 1.95 & 0.24 & 1.44 & 1.89 & 19.46 & 17.89 & 10.30 & 17.99 \\
    \bottomrule
    \end{tabular}%
    }
\end{table}

%% file: Tables/results_real.tex
\begin{table*}[t]
\footnotesize
\centering
\setlength{\tabcolsep}{2pt}
\renewcommand{\arraystretch}{1.15}
\caption{Dynamics prediction errors on the real-world dataset (centimeters; MSE in cm$^2$). All baselines were trained on the proposed
simulation dataset. ``-first'' rows use a single sample; PointZero-FM and PointZero-JiT
report all metrics of the sample with the lowest MDE out of 10 (best-of-10).
\textbf{Bold} marks the best result in each column, and \underline{underlining} indicates
the second-best result.}
\label{tab:results_real}
\resizebox{\linewidth}{!}{%
\begin{tabular}{@{}l cccc cccc cccc@{}}
\toprule
\multirow{2}{*}{Method}
& \multicolumn{4}{c}{Deformable}
& \multicolumn{4}{c}{Articulated}
& \multicolumn{4}{c}{Rigid} \\
\cmidrule(lr){2-5}\cmidrule(lr){6-9}\cmidrule(lr){10-13}
& MDE $\downarrow$ & MSE $\downarrow$ & CD $\downarrow$ & EMD $\downarrow$
  & MDE $\downarrow$ & MSE $\downarrow$ & CD $\downarrow$ & EMD $\downarrow$
  & MDE $\downarrow$ & MSE $\downarrow$ & CD $\downarrow$ & EMD $\downarrow$ \\
\midrule
GBND & 6.179 & 81.114 & 4.039 & 6.069 & 3.051 & 24.148 & 2.089 & 2.912 & 4.379 & 34.418 & 2.085 & 4.307 \\
ParticleFormer & 6.156 & 80.928 & 4.022 & 6.044 & 3.026 & 24.009 & 2.078 & 2.887 & 4.367 & 34.332 & 2.081 & 4.296 \\
PGND & 3.757 & 27.054 & 2.244 & 3.646 & 2.828 & 20.137 & 1.702 & 2.666 & 2.650 & 13.210 & 1.212 & 2.546 \\
PTv3 & 4.705 & 35.109 & 2.975 & 4.606 & 4.430 & 43.665 & 2.645 & 4.284 & 2.487 & \textbf{11.604} & 1.108 & 2.371 \\
\modelname{}-FM-first & 3.739 & 28.765 & 2.293 & 3.642 & 2.046 & 10.152 & 1.168 & 1.883 & 2.711 & 15.697 & 1.216 & 2.542 \\
\modelname{}-FM & \underline{2.923} & \textbf{17.385} & \underline{1.852} & \textbf{2.827} & \textbf{1.720} & \textbf{8.096} & \textbf{1.031} & \textbf{1.580} & \textbf{2.322} & 12.484 & \textbf{1.078} & \textbf{2.182} \\
\modelname{}-Regression & 3.643 & 25.258 & 2.416 & 3.584 & 2.192 & 11.227 & 1.322 & 2.060 & 2.654 & 14.647 & 1.133 & 2.558 \\
\modelname{}-JiT-first & 3.706 & 32.545 & 2.294 & 3.636 & 2.317 & 12.608 & 1.338 & 2.166 & 2.815 & 15.969 & 1.276 & 2.656 \\
\modelname{}-JiT & \textbf{2.915} & \underline{19.162} & \textbf{1.837} & \underline{2.846} & \underline{1.914} & \underline{9.670} & \underline{1.156} & \underline{1.767} & \underline{2.398} & \underline{12.376} & \underline{1.103} & \underline{2.238} \\
\bottomrule
\end{tabular}%
}
\end{table*}

%% file: Tables/pgnd_zeroshot.tex
\begin{table*}[t]
\footnotesize
\centering
\setlength{\tabcolsep}{2.0pt}
\renewcommand{\arraystretch}{1.08}
\caption{Evaluation of different 3D dynamics architectures on zero-shot predictions on
the PGND benchmark~\cite{zhang2024particle}. All methods were trained on synthetic data
from the \modelname{} dataset, and evaluated on unseen PGND scenes. Errors are in
centimeters. \textbf{Bold} marks the best result in each column.}
\label{tab:pgnd_zeroshot}
\resizebox{\textwidth}{!}{%
\begin{tabular}{@{}l *{18}{c}@{}}
\toprule
\multirow{2}{*}{Method}
& \multicolumn{3}{c}{Box}
& \multicolumn{3}{c}{Paperbag}
& \multicolumn{3}{c}{Cloth}
& \multicolumn{3}{c}{Rope}
& \multicolumn{3}{c}{Sloth}
& \multicolumn{3}{c@{}}{Bread} \\
\cmidrule(lr){2-4}\cmidrule(lr){5-7}\cmidrule(lr){8-10}\cmidrule(lr){11-13}\cmidrule(lr){14-16}\cmidrule(l){17-19}
& \mde & \cd & \emd & \mde & \cd & \emd & \mde & \cd & \emd
& \mde & \cd & \emd & \mde & \cd & \emd & \mde & \cd & \emd \\
\midrule
PGND-ZS & 3.1 & 2.0 & 3.0 & 2.9 & 1.7 & 2.7 & 5.9 & 3.0 & 5.6 & 8.6 & 5.1 & 8.5 & 8.3 & 3.1 & 7.5 & 2.9 & 1.3 & 2.8 \\
GBND-ZS & 2.5 & 1.6 & 2.4 & 3.7 & 2.2 & 3.6 & 8.9 & 4.5 & 8.7 & 13.0 & 8.8 & 12.9 & 11.1 & 4.8 & 10.3 & 3.3 & 1.6 & 3.3 \\
ParticleFormer-ZS & 2.5 & 1.6 & 2.4 & 3.7 & 2.1 & 3.6 & 8.9 & 4.5 & 8.7 & 13.0 & 8.8 & 12.9 & 11.1 & 4.8 & 10.3 & 3.3 & 1.6 & 3.3 \\
PTv3-ZS & 3.4 & 2.0 & 3.3 & 3.9 & 2.2 & 3.7 & 7.2 & 3.6 & 6.9 & 7.7 & 4.9 & 7.6 & 6.3 & 2.5 & 5.2 & 2.6 & 1.2 & 2.5 \\
\modelname{}-FM-ZS & \textbf{2.2} & \textbf{1.4} & \textbf{2.1} & \textbf{2.7} & \textbf{1.6} & \textbf{2.5} & 5.1 & \underline{2.5} & 4.9 & \textbf{5.4} & \textbf{3.5} & \textbf{5.3} & \textbf{5.6} & \textbf{2.3} & \textbf{4.9} & \textbf{1.9} & \textbf{0.9} & \textbf{1.8} \\
\modelname{}-Regression-ZS & 2.5 & 1.6 & \underline{2.3} & \underline{2.8} & \underline{1.7} & \underline{2.7} & \underline{4.9} & \underline{2.5} & \underline{4.7} & 5.9 & \underline{3.6} & \underline{5.8} & 6.3 & \underline{2.5} & \underline{5.5} & 2.4 & 1.1 & 2.4 \\
\modelname{}-JiT-ZS & \underline{2.3} & \underline{1.5} & \textbf{2.1} & \textbf{2.7} & \textbf{1.6} & \textbf{2.5} & \textbf{4.7} & \textbf{2.4} & \textbf{4.4} & \underline{5.8} & 3.8 & \underline{5.8} & \underline{6.1} & 2.6 & \underline{5.5} & \underline{2.1} & \underline{1.0} & \underline{2.0} \\
\bottomrule
\end{tabular}%
}
\end{table*}